\pdfoutput=1

\documentclass[11pt]{article}

\usepackage{acl}

\usepackage{times}
\usepackage{amsthm}
\usepackage{latexsym}
\usepackage{graphicx}
\usepackage{color}
\usepackage{caption}
\usepackage{subcaption}
\usepackage{bbm}
\usepackage{url}
\usepackage{amssymb}
\usepackage{amsmath}
\usepackage{multirow}
\usepackage{booktabs}
\usepackage{algorithm}
\usepackage{algpseudocode}
\usepackage{wrapfig}

\title{A Survey of Early Exit Deep Neural Networks in NLP}

\author{Divya Jyoti Bajpai and Manjesh Kumar Hanawal\\
   Department of IEOR, IIT Bombay \\
  \texttt{\{divyajyoti.bajpai, mhanawal\}@iitb.ac.in}}

\begin{document}
\maketitle
\begin{abstract}
Deep Neural Networks (DNNs) have grown increasingly large in size to achieve state-of-the-art performance across a wide range of tasks. However, their high computational requirements make them less suitable for resource-constrained applications. Also, real-world datasets often consist of a mixture of easy and complex samples, necessitating adaptive inference mechanisms that account for sample difficulty. Early exit strategies offer a promising solution by enabling adaptive inference, where simpler samples are classified using the initial layers of the DNN, thereby accelerating the overall inference process. By attaching classifiers at different layers, early exit methods not only reduce inference latency but also improve the model’s robustness against adversarial attacks. This paper presents a comprehensive survey of early exit methods and their applications in NLP. 
\end{abstract}

\section{Introduction}
Deep Neural Networks (DNNs) such as BERT \cite{devlin2018bert}, GPT \cite{radford2019language}, XLNet \cite{yang2019xlnet}, ALBERT \cite{lan2019albert}, ViT \cite{alexey2020image}, BLIP-2 \cite{li2023blip}, Llama \cite{touvron2023llama} etc., have expanded significantly in size, achieving significant improvements in various Image and Natural Language Processing (NLP) tasks.
These models leverage large-scale pre-training on unlabeled data, followed by fine-tuning on labeled datasets to deliver state-of-the-art performance. The large size of these DNNs introduces several challenges in deployment. The first major issue is deploying them on resource-constrained devices such as mobile phones, edge devices, and IoT platforms to maintain their high performance. The second issue is `overthinking', where DNNs continue processing even when shallow layers could produce correct inferences for easier samples as shown in \cite{kaya2019shallow, michel2019sixteen, zhou2020bert}. This unnecessary deep processing can overfit irrelevant features, resulting in poor generalization and wasted computation. Additionally, overthinking contributes to the models' susceptibility to adversarial attacks \cite{zhou2020bert}.

To address these issues, recent research has focused on accelerating DNN inference and making their implementation feasible for limited-resource environments. Techniques like pruning \cite{fan2019reducing, michel2019sixteen}, quantization \cite{zhang2020ternarybert, bai2020binarybert, kim2021bert} and knowledge distillation \cite{sanh2019distilbert, jiao2019tinybert} have been employed to reduce the size of DNNs. These compression methods decrease the model size but often sacrifice the optimal performance and versatility of the original networks. These methods use the same processing on each sample without any adpation, which makes them static, leading to suboptimal performance and inefficient usage of resources. Real-world tasks consist of samples with varying levels of complexity, hence they do not need the same computational effort. This variability calls for input-adaptive inference methods that tailor the computational effort to the complexity of each input.

\begin{figure}
    \centering
    \includegraphics[width=0.71\linewidth]{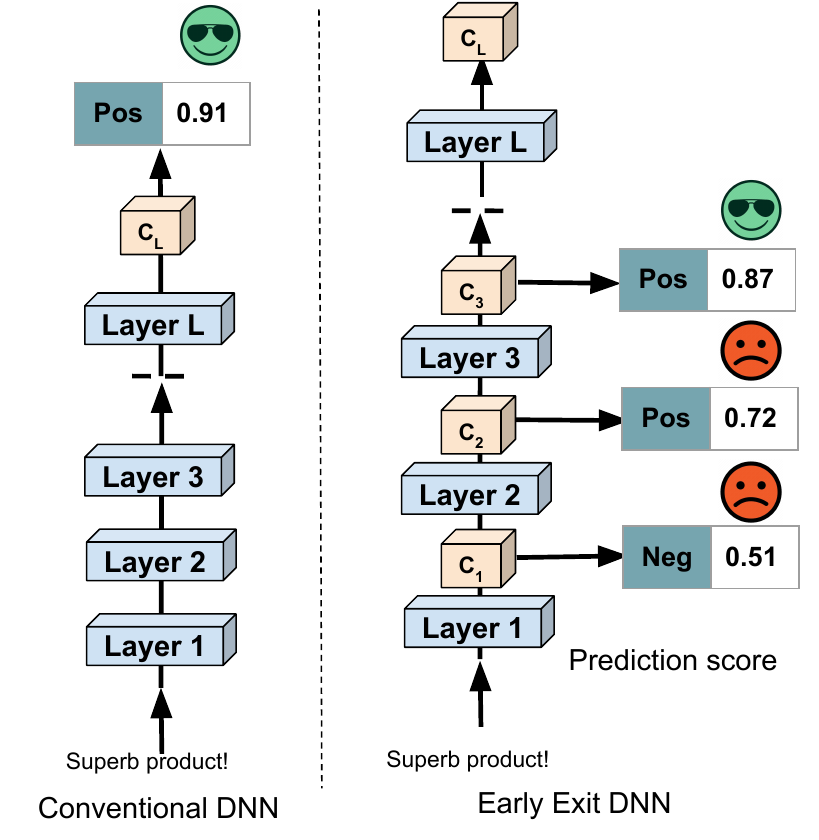}
    \caption{Difference between the DNN and EEDNN.}
    \label{fig:DNNvsEEDNN}
\end{figure}

Early Exit (EE) \cite{teerapittayanon2016branchynet} methods have emerged as a state-of-the-art input-adaptive approach to address the challenges of overthinking and latency in DNN inference. These methods incorporate intermediate classifiers at several layers within the DNN, allowing inference to occur at multiple stages. The inference process halts once the model reaches a sufficient level of confidence in its prediction, enabling dynamic, `anywhere' predictions. Samples that achieve high prediction confidence at the at shallower layers exit early, while only more complex samples are processed deeper into the network \cite{xu2022survey}. In Figure \ref{fig:DNNvsEEDNN}, we show the conventional DNN and EEDNN where conventional DNN exits the sample only at the final layer, while the EEDNN infers the sample at the 3rd layer as it gains sufficient confidence there. Anywhere classification allows these models to be partitioned and utilized for edge-cloud co-inference setup where part of the DNN is deployed on the edge and full-fledged DNN on the cloud.

The EE methods have been widely popular in NLP tasks, where they are applied to Large-Language Models (LLMs) and Vision-Language Models (VLMs). Also, there are very few systematic surveys on early exit DNNs. \citet{matsubara2022split} touches upon the EE framework as the application for edge-cloud co-inference setup. \citet{han2021dynamic} reviewed the complete area of dynamic neural networks. Being, one subset of the dynamic neural networks, it only touches upon the EE networks. \citet{rahmath2024early} reviews EEDNNs mostly on image tasks and briefly touches upon the NLP methods.

Since EE methods have been widely adopted for NLP tasks, a comprehensive survey of EEDNNs for NLP is lacking. This gap motivates us to undertake this survey. The aim of this survey is to (1) provide a thorough overview and new insights for researchers interested in early exit methods for NLP; (2) highlight the interconnections between different subareas, thereby minimizing redundancy and the risk of reinventing the wheel; and (3) summarize key challenges and outline potential directions for future research in this evolving field.

\begin{figure}
    \centering
    \includegraphics[width=0.85\linewidth]{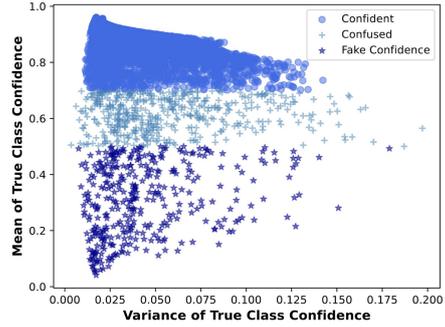}
    \caption{The figure shows the average of the confidence values over the true class across all the layers for the SST-2 dataset.}
    \label{fig:scatter_plot}
\end{figure}

\section{Advantages of EEDNNs}

EE methods offer several advantages over static models by dynamically adjusting computation based on the complexity of incoming samples. The key benefits of EE models are outlined below:

\textbf{1) Faster Inference:} EE models come with additional side branches (exits) attached to the DNN. A significant advantage of EE models is their ability to allocate computational resources selectively at inference time, activating only relevant sub-networks based on the input sample due to attached exits. This results in faster inference, as computational effort is minimized for simpler, easier-to-recognize samples.

\textbf{2) Input-Adaptiveness:} 
EE models adapt computational effort based on the complexity of incoming samples, using less power for easier samples without compromising accuracy. Figure \ref{fig:scatter_plot} illustrates this by plotting average confidence values on the true class across intermediate exits and the final layer of the BERT models with EE at every layer. Approximately $80\%$ of samples, labeled as `confident,' exhibit high confidence and are predicted in the initial layers. `Confused' samples show fluctuating confidence across classes, indicating model uncertainty. Finally, `fake confidence' samples fall outside the model's scope, where the model incorrectly becomes confident about the wrong class, leading to mispredictions.

\textbf{3) Generality:} EE methods are versatile and can be applied to a wide range of tasks, including image classification, object detection, natural language processing, text generation, and image captioning, often with minimal modifications to the model design. This generalizability allows EE methods developed for one task to be easily adapted to others.

\textbf{4) Interpretability:} EE models enhance the interpretability of DNNs by providing insights into the decision-making process at each stage of the network. By allowing users to observe which samples exit early and which proceed to the deeper layers. These models offer a clearer understanding of how the network differentiates between simpler and more complex samples, facilitating a better understanding of the data being processed. For instance, Figure \ref{fig:scatter_plot} provides a deeper insight into the hardness of incoming samples and can help detect OOD samples from a dataset that is out of the model's scope. 

\textbf{5) Robustness:} EE models demonstrate increased robustness against adversarial attacks compared to traditional DNNs. The use of multiple intermediate classifiers creates an ensemble effect, where the impact of noise or adversarial perturbations is mitigated by leveraging predictions from different layers, resulting in more reliable and confident final outputs \cite{zhou2020bert}.

\textbf{6) Distributed Inference:} EE models offer anytime prediction by attaching intermediate classifiers, making them well-suited for varying computational budgets and hardware constraints. This adaptability allows EE models to operate effectively across different hardware platforms and dynamic environments, making them particularly valuable in distributed computing setups \cite{teerapittayanon2017distributed}. It could be easily adapted to various mobile-edge, edge-cloud or mobile-edge-cloud co-inference setups.

\textbf{7) Mitigates Overthinking:} EEs also solve the overthinking issue in DNNs by not forcing a sample to pass through deeper layers even when the sample has gained enough confidence in the initial layers. Sometimes excessive processing of easy samples deeper into the backbone may lead to wrong prediction due to irrelevant feature extraction. Mitigating this not only improves accuracy but also reduces wasteful computation.

Other than these EEs, they also help reduce overfitting, where the interaction between different side branches acts as a regularizer for the model. This solves the vanishing gradient problem by giving the gradient signal from the initial layer that is less prone to vanishing gradient issues.

These properties make Early Exit methods a powerful tool for deploying DNNs in resource-constrained environments and diverse application areas, where efficiency, adaptability, and robustness are critical. They have been widely adopted in various fields such as image classification \cite{teerapittayanon2016branchynet, huang2017multi, laskaridis2020spinn,dai2020epnet, wang2020accelerating, fang2020flexdnn, li2019edge, phuong2019distillation, li2019improved, wolczyk2021zero, khademsohiselfxit}, NLP tasks \cite{bapna2020controlling, elbayad2019depth, liu2021elasticbert, balagansky2022palbert, xin2021berxit, sun2022simple, gao2023pf, bajpai2024ceebert, miao2024efficient}, image captioning \cite{fei2022deecap, tang2023you, miao2024efficient, bajpai2024capeen} etc.

\subsection{Areas of research}
While Early Exit (EE) methods effectively address the above-mentioned issues in DNN inference, they require careful design choices regarding the confidence metric, training strategies and exit criteria. Training EE-based DNNs (EEDNNs) is inherently a multi-objective problem since each intermediate classifier aims to optimize its performance. The decision to exit at a particular layer is based on the intermediate classifier being confident and is governed by a confidence metric that must exceed a predefined threshold. This threshold setting is critical to the inference process, as a higher threshold allows for more accurate predictions at deeper layers but may also increase latency, while a lower threshold does the opposite.

Research on EEDNNs has primarily focused on improving specific aspects, as summarized below:

\textbf{1) Exiting Criteria:} A key area of research involves the choice of confidence metrics and threshold settings tailored to specific tasks. This includes strategies for leveraging the outputs of multiple intermediate classifiers to achieve a better estimate of the true label and setting thresholds that balance the trade-off between accuracy and efficiency \cite{zhou2020bert, balagansky2022palbert, zhang2022pcee, xin2020deebert, bajpai2024ceebert}.

\textbf{2) Training Strategies:} The training of exit classifiers at multiple layers poses a multi-objective optimization problem. The task of each intermediate layer in the EEDNN has two objectives: 1) Provide hidden representations such that the exit classifier loss is minimized. 2) Hidden representations should be such that the final layer accuracy is also not compromised.

Various training approaches have been investigated, such as joint optimization of all exits or separate optimization of each exit and the backbone. Some works also distil the knowledge from deeper layers to initial layers for better learning of the intermediate classifiers. Additionally, attaching classifiers at multiple layers introduces more parameters to the model, which raises the question of how to strategically place these exits across the network to avoid excessive model size, particularly for very large models \cite{ zhu2021leebert, zhou2020bert, wang2019dynexit, xin2021berxit}.

\textbf{3) Generalization of EEDNNs:} While large DNNs generally exhibit strong generalization capabilities, EEDNNs can inherit these properties, but task-specific confidence metrics and thresholds often constrain their generalization. As the domain of the input data changes, the distribution of confidence scores at the exits can also shift, which requires addressing such concerns \cite{bajpai2024ceebert, bajpai2024dadee}.

\textbf{4) Handling Complex Tasks:} For more complex tasks, such as text generation, EEDNNs tend to suffer a greater performance drop. This is because the earlier layers typically capture only syntactic information, while deeper layers are required to extract semantic meaning. A challenge remains in how to equip the initial layers of EEDNNs with the higher-level information typically found in deeper layers of the network \cite{fei2022deecap, bajpai2024capeen}.



\section{Foundation of Early Exit DNNs}
\begin{figure}
    \centering
    \includegraphics[scale = 0.399]{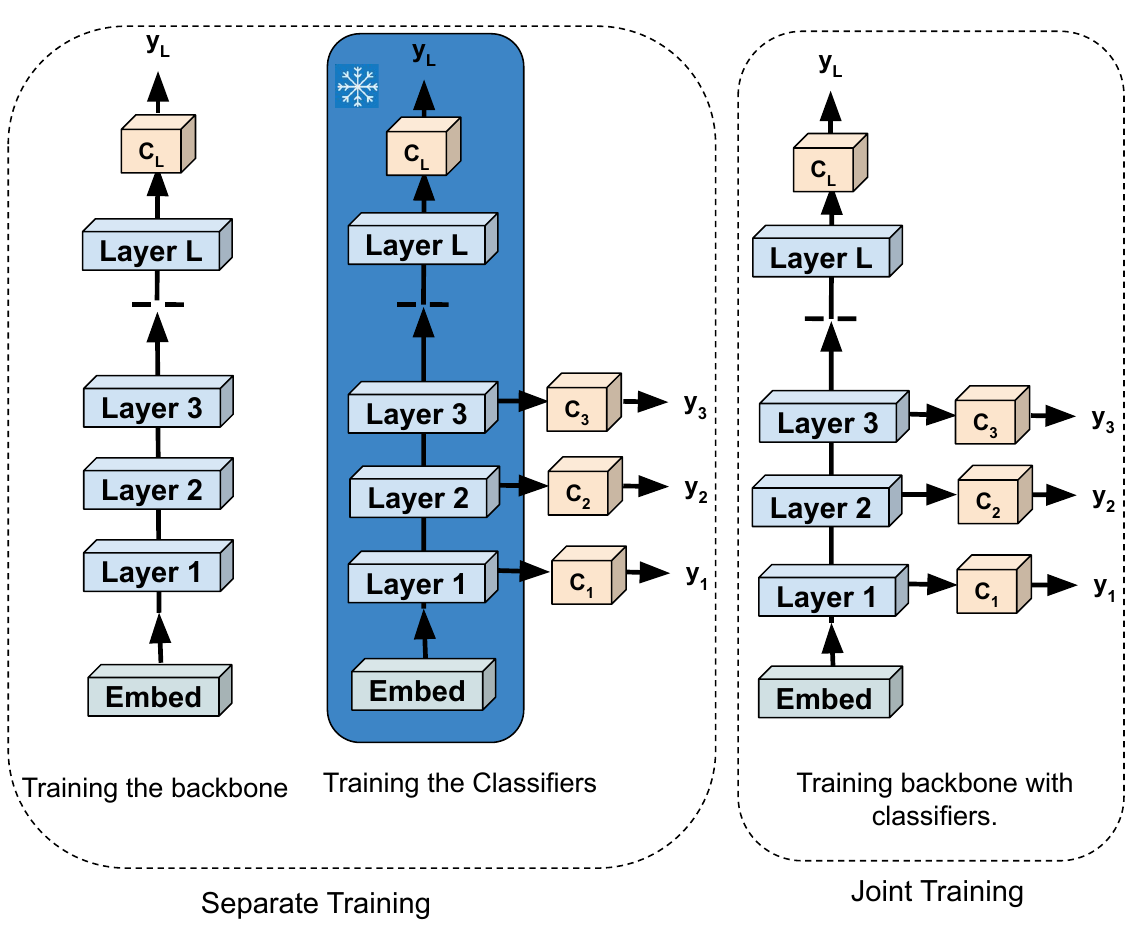}
    \caption{Separate training vs Joint Training}
    \label{fig:Separate_vs_joint}
\end{figure}
EEDNNs belong to a class of dynamic neural networks that adaptively adjust the inference process, by selectively using a subpart of the model based on input sample complexity. In this section, we outline the general framework of EEDNN models: their typical training and inference procedures.

\begin{figure*}
    \centering
    \includegraphics[scale = 0.399]{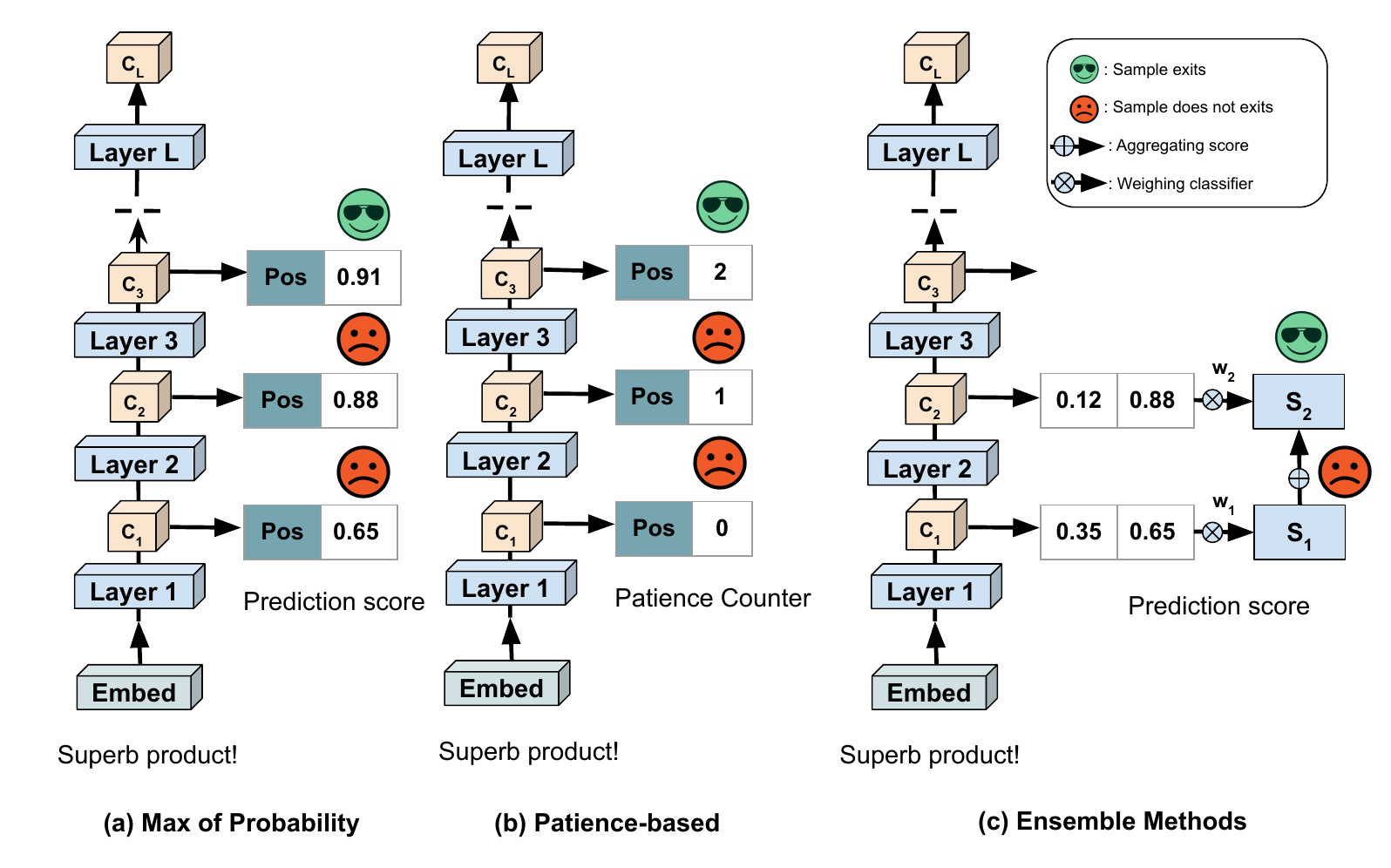}
    \caption{Inference methods: 1) Max Probability: confidence is the maximum output of an individual classifier. 2) Patience-based: relies on prediction consistency between classifiers. 3) Ensemble: aggregates weighted results from multiple classifiers.}
    \label{fig:inference}
\end{figure*}

\subsection{Setup}
To construct an EEDNN, classifiers are integrated at intermediate layers to map the hidden representations of the backbone network to output probabilities. These additional classifiers not only provide regularization to the main network but also offer more direct gradient signals for backpropagation, particularly from shallower layers.

In designing an EEDNN, several key factors must be considered: (1) the training strategy for classifiers at all intermediate layers; (2) the architecture of the classifiers, including their size, depth, and complexity (e.g., a single linear layer \cite{xin2020deebert}, multiple fully connected layers \cite{fei2022deecap} and combination of self-attention and fully connected layers; (3) the exit criteria for each classifier and the associated computational cost; and (4) the optimal placement of exit points.

\subsection{Training methods}
\textbf{Separate Training:}
Methods such as \citet{xin2020deebert, bajpai2024capeen} perform separate training as detailed below and in Figure \ref{fig:Separate_vs_joint}.
Let us consider that there are $N$ layers in the backbone. We also consider that $\mathcal{D}$ represents the distribution of the dataset with a label class $\mathcal{C}$ used for the backbone training.  
For fine-tuning the backbone, the loss function for $i$th exit is written as:
\begin{equation}
    \mathcal{L}_i(\theta) = \mathcal{L}_{CE}(f_i(x, \theta), y)
\end{equation}

Here, $f_i(x, \theta)$ is the output of the classifier attached at the $i$th exit, $\theta$ denotes the collection of all the parameters, $\mathcal{L}_{CE}$ is the cross-entropy loss and $(x, y)\sim \mathcal{D}$. 

In separate training, the network undergoes fine-tuning in two stages:

1) The first stage involves updating the embedding layer, all transformer layers, and the final classifier, with the loss function being solely $\mathcal{L}_{N}$. This is just standard backbone fine-tuning.

2) In the second stage, the parameters fine-tuned in the first stage are frozen, and only the remaining components, excluding the final classifier, are updated. Here, the loss function is $\sum_{i=1}^{N-1}w_i\mathcal{L}_i$. This approach ensures that the backbone parameters remain fixed to preserve their optimal quality; otherwise, the transformer layers might no longer be optimized exclusively for the final layer, which generally leads to a decline in its performance.

\textbf{Joint Training:}
Methods such as \citet{zhou2020bert, bajpai2024ceebert} perform Joint Training where instead of first finetuning the backbone and freezing its weights, the complete backbone is simultaneously optimized (see Figure \ref{fig:Separate_vs_joint}). Hence the loss function is:
$\mathcal{L} = \sum_{i=1}^{N}w_i\mathcal{L}_i$. This method simultaneously finetunes the backbone and learns the classifier weights.

The weights $w_i$ in both the separate and joint training are the weights provided based on the cost associated with each exit classifier. Most of the methods replace $w_i = i$ with a justification that more emphasis should be given to deeper layers. However, DynExit \cite{wang2019dynexit} proposes $w_i$ to be trainable parameters and use $\sigma(w_i)$ instead of $w_i$ where $\sigma$ is the sigmoid function. After this step, the backbone is ready for inference. 


\textbf{Other methods:}
Some methods use a combination of the existing methods such as BERxiT \cite{xin2021berxit} uses the alternate training where in one iteration the backbone weights are optimized and in the next step the exit weights are optimized. As the exits have two objectives, the motivation for using this method is to have a good balance between the two objectives.

Other than these methods some works additionally use knowledge distillation between the layers \cite{zhu2021leebert, geng2021romebert} or distillation from the final layer to the other intermediate classifiers \cite{bajpai2024capeen}. 
\subsection{Defining confidence}
After training the backbone, it is necessary to define the confidence of the exit classifiers. This subsection details different measures of confidence for deciding to exit. 

\textbf{Individual confidence-based:}
Let $\hat{P}_i(c)$ represent the estimated probability that input $x$ belongs to class $c \in \mathcal{C}$, and let $C_i$ denote the confidence in this estimate for the $i$th exit. CeeBERT \cite{bajpai2024ceebert} defines confidence as the maximum estimated probability across all classes, i.e., $C_i := \max_{c \in \mathcal{C}} \hat{P}_i(c)$. In contrast, DeeBERT \cite{xin2020deebert} and ElasticBERT \cite{liu2021elasticbert} use the entropy of the $\hat{P}_i(c)$ as the confidence score. Note that these methods only use the output from a single classifier.

\textbf{Patience-based:}
PABEE \cite{zhou2020bert} takes a different approach by defining confidence based on prediction consistency across multiple exit classifiers. If predictions from several consecutive classifiers remain consistent, the sample is inferred. LeeBERT \cite{zhu2021leebert} also utilizes patience-based exiting similar to PABEE. The advantage of this method is that it reduces the chances of adversarial attacks as its predictions are based on multiple classifier's output.

\textbf{Distribution-based:}
In this category, works like PALBERT \cite{balagansky2022palbert} introduce the Q-exit strategy, where a distribution $p(i|x)$ is learned over exit classifiers, representing the probability that a sample exits at the $i$th layer. A sample exits the backbone once the cumulative distribution function (CDF) exceeds a predetermined threshold. JEI-DNN \cite{chataoui2023jointly} learns the distribution over the exit layers using joint optimization without requiring additional training. The major advantage of this method is it does not require to verify the confidence at every exit instead for every incoming sample, an intermediate exit is assigned and it is directly inferred at that exit.

\textbf{Similarity-based:}
MuE \cite{tang2023you} model decides upon exiting based on the similarity score of the consecutive layers. At every layer, the similarity of hidden representations with the previous layer is calculated and if it is less than the given threshold, the sample exits the backbone. The motivation for this method comes from the fact that the hidden representations saturate once sufficient features are extracted. The advantage of this method is that it reduces the need for checking the confidence values after processing through the exit instead it can decide to exit based on similarity reducing computational demands.

\textbf{Ensemble methods:}
Methods such as ZTW \cite{wolczyk2021zero} use ensemble-based exiting criteria where weights are provided to different classifiers depending on the confidence in the classifier's prediction, a sample is exited from the backbone once the ensemble score exceeds a predefined threshold. Similarly, \citet{sun2021early} uses a majority vote to decide early inference of a sample, if a certain number of classifiers agree on one class, the sample exits the backbone. The advantage of this method is the ensemble of multiple classifiers making predictions more trustworthy.

\textbf{Other methods:}
BERxiT \cite{xin2021berxit} introduces learning-to-exit modules that use a separate network to estimate sample uncertainty rather than traditional confidence measures. HASHEE \cite{sun2022simple} employs a hash-based strategy, assigning exit layers based on sample clustering based on frequency or embedding space, without relying on confidence. \citet{gao2023pf} combine patience and similarity-based methods, exits when consecutive layer similarities fall below a threshold repeatedly. \citet{he2024cosee} uses signal-based exiting, allowing exits to prioritize samples likely to exit under different acceleration scenarios.

\subsection{Choice of thresholds}
The threshold used to decide whether to exit is a crucial part of the EEDNNs. The threshold models the accuracy-efficiency trade-off. The ways to set the thresholds are as follows:

\textbf{Static thresholds:}
Methods such as BranchyNet \cite{teerapittayanon2016branchynet}, PABEE \cite{zhou2020bert}, LeeBERT \cite{zhu2021leebert}, DeeBERT \cite{xin2020deebert}, DeeDiff \cite{tang2023deediff}, FastBERT \cite{liu2020fastbert}, FlexDNN \cite{fang2020flexdnn}, DynExit \cite{wang2019dynexit}, etc. set the threshold based on the best-performing threshold on the validation split of the dataset. Most of the methods focus on maximizing the accuracy of the validation set. These methods apply a static threshold either by greedily choosing the threshold based on accuracy or some combination of accuracy and latency which is not the goal always.

\textbf{Dynamic thresholds:}
Methods such as CeeBERT \cite{bajpai2024ceebert} and UCBEE \cite{pacheco2024ucbee} model the problem of choosing the optimal threshold using a Multi-Armed Bandit (MAB) setup. In their mobile-cloud co-inference setup, the threshold is used to decide if a sample can be inferred locally or should be offloaded to the cloud. CeeBERT \cite{bajpai2024ceebert} on the other hand learns the optimal threshold using Multi-Armed Bandits setup under the case that the test data distribution is different from the training dataset. It defines a reward function for the threshold consisting of both the confidence in prediction and the cost of processing a sample into the backbone. MuE \cite{tang2023you} also uses a dynamic threshold for image captioning tasks where the threshold value decreases with the increasing length of the sentence. MuE claims that the decoder tends to make fewer mistakes as the sentence length gets longer.

\subsection{Inference}
During inference, as an input instance $x$ sequentially passes through layers $1, \ldots, L$, each exit classifier positioned after the intermediate layers produces a class label distribution. The inference process halts at the $i$th exit classifier when the confidence score $C_i$ satisfies $C_i \geq \alpha$, where the definition of $C_i$ is as described in the previous section. If the model does not reach a sufficient confidence level by the final layer, the sample is inferred at the final layer regardless of its confidence score. This mechanism enables early exiting of a sample from the backbone when the confidence condition is met, thus avoiding unnecessary traversal through all layers.

\section{Applications}
In this section, we provide details of the applications of the early exit methods to different NLP domains, such as text classification, natural language inference (NLI), Language Translation, Sequence Labeling and Image captioning tasks.

\subsection{Text classification and NLI tasks}
In most of the NLP tasks, the EE methods only attach a linear classifier in the exit instead of a complex structure as done on the image tasks. DeeBERT \cite{xin2020deebert} first applied EEs to the BERT backbone, it performed a separate training and uses entropy as the confidence metric.
ElasticBERT \cite{liu2021elasticbert} on the other hand performs the training of the BERT backbone from scratch i.e., during pre-training of the BERT backbone, the MLM and SOP heads are attached to every layer instead of just the final layer. Hence after pre-training the backbone has learned weights such the objective is not only to improve the final layer's performance. By pertaining the backbone from scratch with exits, it optimizes the performance of the backbone for EE and final layer.

Some works such as PABEE \cite{zhou2020bert} highlight the overthinking issues in the NLP tasks and also show that these models not only perform faster inference but also make the original model robust to adversarial attacks. Since PABEE proposes patience-based exiting criteria i.e., based on prediction consistency, it does not rely on a single classifier to decide exiting which makes it more robust to the noise in the incoming samples.

BERxiT \cite{xin2021berxit} performs an alternating training strategy where in one iteration the full backbone is optimized and in the next iteration the exits are optimized. The exiting criteria are learned where the decision to exit is taken by a learned single linear layer that outputs uncertainty in prediction. It empirically proves better performance by alternate training and novel learning to exit modules instead of only depending on the confidence of the model. 

Knowledge Distillation (KD) methods, initially used to distil the knowledge of larger models into smaller models have also been explored in early exit models. FastBERT \cite{liu2020fastbert} utilizes this strategy where it first finetunes the BERT backbone and then attaches exits to the backbone. Then the model weights are frozen and only exit weights are trained where additional knowledge distillation loss is applied from the final layer to the student classifiers. LeeBERT \cite{zhu2021leebert} on the other hand, instead of learning from only the final classifier allows knowledge to be distilled within multiple exits. It also uses cross-level optimization by partitioning the training dataset, where the training dataset is optimally split for the backbone and the exit weights training i.e., the dataset used for backbone training is different from the dataset used for exits training. KD loss improves early exit accuracy by providing soft labels with hard labels which improves accuracy as well as efficiency.

Methods such as PALBERT \cite{balagansky2022palbert} and ETFEE \cite{ji2023early} have proposed to alter the exit classifier's configuration where PALBERT extends transformer layers with a Lambda layer that induces a generalized geometric distribution on the of exiting from the $i$th layer equal to $p(i|x) = \lambda_i\prod_{j=1}^{i-1}(1-\lambda_j)$ where $\lambda_i$ is a function of hidden representation at $i$th layer. ETFEE additionally has an adapter whose function is to disentangle the task-specific and universal representations. Also, instead of the classic classifier, an equiangular tight frame (ETF) classifier is added to enhance the classification ability of internal classifiers. Similarly \citet{gao2023pf} utilize the adapter module and perform parameter efficient fine-tuning for the exit classifiers and perform exiting based on the similarity between consecutive hidden layers. In these methods, the exits are computationally expensive but are more accurate as compared to other methods.

\citet{liao2021global} proposed a method that does not use only a single classifier for inference but all the past classifiers using ensemble strategies. It also utilizes the future classifiers that have not been explored by the sample by using an imitation classifier which is a lightweight model with the task of imitating the remaining transformer layers. It has improved the previous state-of-the-art early exiting methods by using all the classifiers and producing an ensemble effect. However, the computational complexity of this method is higher due to additional imitation classifiers that are used to get the information from the deeper layers that might not have been used due to the early exiting.

JEI-DNN \cite{chataoui2023jointly} on the other hand jointly learns a probability distribution along with the classifier weights where it learns a distribution over the set of layers and during inference this distribution is utilized to decide the exiting of the sample from a particular intermediate exit without checking at other exits. This creates a multi-objective problem and all tasks are simultaneously optimized. However, the balance between different tasks needs to be maintained. 

\subsection{Text Summarization}
HASHEE \cite{sun2022simple} has applied early exits for text summarization. Note that text summarization is a more complex task as it involves the generation of text, and hence requires better modelling. The major contribution of HASHEE is it does not require checking the confidence at every layer instead it divides the vocabulary into $n$ buckets where $n$ is the number of exits attached to the backbone. The bucketing could be done based on clustering, frequency and mutual information. Each bucket is assigned one of the exits for inference. For instance, the tokens whose frequency is higher are considered easier and are assigned initial layers and the tokens that rarely appear are assigned deeper layers. In this way, the computational cost is further reduced.

\subsection{Sequence labeling tasks}

\citet{wang2020accelerating} proposed two early exiting strategies for the sequence labeling tasks: 1) Sentence level Early Exit (SENTEE) where complete sentence exits together at one layer. To decide which layer is suitable the uncertainty is defined as the max of uncertainties over each token in the sequence. 2) TOKEE: The main issue of SENTEE is that a sample cannot exit the backbone until each token gets sufficient confidence. To circumvent this TOKEE uses token level exiting i.e., as a token in the sequence gets sufficient confidence, it is not further processed saving the unnecessary computation of taking each token deep into the backbone.

\subsection{Language Translation}

HCN \cite{tsai2022hierarchical} applies early exits to the decoder of transformer models for language translation tasks. It performs separate training and distils final layer knowledge to the exits using knowledge distillation loss The main issue faced was the size of the exits has increasingly grown for the translation tasks. To reduce the size of exits, HCN reduces the vocab size for the shallower layers and makes them learn about the specific token by not adding up the loss of those tokens that are planned to be removed from the vocab size. The choice of the token used for different exits is made in a hierarchal way where top-$ki$ samples were kept for $i$th exit based on their frequency in vocab, where $k$ is some constant. This significantly reduces the exit classifier size further reducing the computational complexity of the model.

\subsection{Vision-language tasks}
Extending early exit methods to vision-language tasks presents unique challenges: 1) Shallow layers primarily capture syntactic information, while deeper layers encode semantic relations, making initial exits lack semantic fusion capabilities. 2) Image captioning models involve a large number of output classes equal to the vocabulary size, resulting in significant parameter overhead when adding classifiers to multiple exits.

DeeCap \cite{fei2022deecap} addresses performance degradation due to missing high-level features by employing lightweight imitation-learning-based networks. An MLP mimics deeper transformer layers using intermediate hidden representations by outputting similar hidden representations as the original transformer backbone, mitigating the lack of high-level features. However, the computational complexity of this method is quite high as the imitation network architecture adds to the latency of the model.

MuE \cite{tang2023you} introduces a similarity-based exit criterion, assuming minimal changes in hidden representations between layers for confident samples. Exits occur when the similarity score between consecutive layers falls below a predefined threshold. Unlike other methods limited to decoders, MuE extends early exiting to the encoder by halting feature extraction when the threshold is met, passing the representations directly to the decoder. The extension to the encoder also reduces the inference time in encoder-decoder models. As the halting process does not depend on the classifier's confidence, it further reduces the inference time for performing inference at every exit.

DEED \cite{tang2023deed} uses adapter modules between exit classifiers and decoder layers to minimize information loss in shallow layers. It standardizes intermediate classifiers across exits and combines final layer loss with the average loss from all exits to preserve backbone optimality.

CapEEN \cite{bajpai2024capeen} introduces a two-step training process: training the backbone without exits, then freezing its weights while training exits using cross-entropy and knowledge distillation losses. Its variant, A-CapEEN, leverages Multi-Armed Bandits to dynamically adjust exit thresholds during inference, adapting to image noise.



\section{Domain Generalization in EE Models}
Large-scale DNNs have strong generalization capabilities across domains with similar tasks \cite{wang2023bert} i.e., if a DNN model is trained on one domain (source domain) say movie reviews, then it performs well when it is tested on other domains (target domain) such as electronic product reviews. However, even when the underlying task is the same, there is a performance drop due to the change in the semantic structure of the reviews of the different domains.

This property of better generalization to various domains is also inherited by EEDNNs as they are extensions of the DNNs. However, note that EEDNNs highly depend on the exit confidence values and the threshold is set based on that using the validation split of the source dataset. However, the confidence distribution at the exits changes due to the change in the domain of the dataset. This change in confidence distribution impacts the trade-off between accuracy and efficiency. It necessitates the requirement of either adapting the threshold value according to the target domain or forcing the backbone to provide domain-invariant features to the classifiers such that the confidence distribution at the exits is not changed. The existing two types of methods are detailed below.

\textbf{Threshold-based adaptation:}
CeeBERT \cite{bajpai2024ceebert} is the first work that tries to solve the issue of domain adaptation in EEDNNs by adapting the threshold based on the unknown domain. Since during the inference phase data arrives in an online and unsupervised manner, hence the problem is to find the optimal threshold when the data arrives in an online and unsupervised manner.

CeeBERT models this problem as a multi-armed bandit setup, where the action set is the set of thresholds. It defines the reward function as the combination of the confidence of the classifier and the latency incurred to get the prediction from the classifier. The reward function is defined such that it increases with an increase in confidence and decreases with an increase in latency. The objective is to maximize the reward function which in turn maximizes confidence over a sample while minimizing the latency incurred. 
Since the confidence distribution is unknown and depends on the target domain, CeeBERT uses the UCB algorithm to solve the problem of finding the optimal threshold. UCB algorithm uses exploration-exploitation principles to identify the best action (threshold).


\textbf{Feature-based adaptation:}
Threshold-based domain adaptation only tunes the threshold based on the new domain. 
DAdEE \cite{bajpai2024dadee} proposes a GAN-based framework to learn domain-invariant features across all the layers. It has a three-step procedure: 1) \textbf{Supervised training:} First a backbone with attached exits is trained on the source domain with labels that perform well on the source dataset. 2) \textbf{Unsupervised domain adaptation:} In this step, the domain adaptation takes place in a GAN-based setup. At every layer, DAdEE attaches a discriminator with a task to discriminate if a feature representation is from the source domain or target domain. All the layers have a task to generate representations such that the discriminator can be fooled and cannot distinguish between the source and target domain. Knowledge distillation is used to reduce the impact of mode collapse, which is common in GANs. 3) \textbf{Inference:} Finally, the third step involves performing inference using the same classifiers as trained on the source domain. Since the new model now generates representations that cannot be distinguished between source and target domain, it justifies the use of similar classifiers.  

\section{Further Applications}

\textbf{OOD Detection:}
Early Exit methods have also been used for OOD detection by \citet{zhou2023two} where the task is to determine the out-of-distribution sample where the original backbone was trained on the in-domain samples. The training loss is modified and added with a relative loss that assesses the interdependency between exits.

During inference, the OOD sample is identified as a sample that has not gained a sufficient number of votes from the classifiers. A sample is first passed through the backbone and if the majority vote of the classifiers reaches a certain threshold then the sample is early inferred else, it is labeled as an OOD sample. 

\textbf{Reinforcement learning:}\label{sec: ZTW}
ZTW \cite{wolczyk2021zero} applies the early exit framework to the Reinforcement Learning algorithm to accelerate their inference time. It implements the idea of cascaded connections by adding skip connections that combine the output of $m$th layer of the model with $(m-1)$th layer classifier output and passes it to the $m$th classifier. This makes the model aware of the previous classifier's output and helps the model to provide more confident results. ZTW experiments with Q\textsuperscript{*}-BERT and Pong, two popular Atari 2600 environments.


\textbf{Self-speculative decoding:}
Speculative decoding is a method used to reduce the latency issues in autoregressive decoding tasks. In this method, two models are used, where a smaller \textit{draft} model is used to generate the tokens in an autoregressive manner and then a larger model \textit{verifies} the output of the draft model in a non-autoregressive manner saving lot of computation without losing accuracy.

Recently LayerSkip \cite{elhoushi2024layer} and Draft \& Verify \cite{zhang2023draft} combine early exits with speculative decoding and name it as \textit{self-speculative decoding}. In this setup, the draft model is replaced by some initial layers of the large model. The early exit point is attached at a chosen layer and then the tokens are generated in an autoregressive manner and the tokens are verified using the final layer of the model.


\textbf{Distributed Inference:}
Early exit (EE) methods optimize distributed inference across mobile, edge, and cloud devices by enabling samples to exit on different devices based on confidence, reducing offloading costs. DDNN \cite{teerapittayanon2017distributed} pioneered this approach, but three key challenges arise:
1) Optimal partitioning layer: SplitEE \cite{bajpai2023splitee, bajpai2024splitee} address this using a Multi-Armed Bandit (MAB) framework.
2) Optimal threshold: UCBEE \cite{pacheco2024ucbee} tackles threshold selection as an MAB problem, optimizing over a predefined set.
3) DNN during outages: UEEUCB \cite{hanawal2022unsupervised} optimize exit points with MABs, targeting image and NLP tasks, respectively. DEE \cite{ju2021dynamic} enhances robustness in dynamic conditions using contextual bandits to handle distributional shifts.

\section{Future Directions}
In this section, we list some of the possible future research directions. 

\subsection{Exit placement and size}
For smaller models like BERT and ALBERT, exit classifiers can be placed after every layer due to the limited number of layers. However, for larger models such as LLAMA and OPT, this approach significantly increases parameters. For instance, adding a classifier to each layer of OPT\textsubscript{2.7B}, with a hidden size of 2560 and a vocabulary size $\mathcal{V}$, results in $130M$ parameters per classifier. With 32 layers, this totals $4B$ parameters, exceeding the model size itself.

Additionally, placing more exits in initial layers improves efficiency but can lead to higher performance degradation, while exits in deeper layers reduce performance drops but compromise efficiency. To balance these trade-offs, exits should be strategically placed at intervals, as consecutive layers often yield minimal additional information, necessitating careful selection of layers for exit attachment.

\subsection{Risk in EEDNNs}
Similar to DNNs, EEDNNS are also prone to the risk of getting the wrong prediction. Note that the EEDNNs are even at more risk as there are multiple classifiers that can get wrong predictions. This issue is brought up in Fast yet Safe \cite{jazbec2024fast} paper where they show that the threshold used for early exiting could also be used to minimize the risk. However, it has very less insights on if the model gains fake confidence over the wrong class and gets predicted early. A thoughtful consideration of this issue is necessary. 

\subsection{Overconfidence}
In Figure \ref{fig:scatter_plot}, we plot the average confidence values of the exit classifiers across the backbone on the true label of the incoming sample. The dataset used is the SST-2 dataset with a task of sentiment classification. We can observe that there are samples marked as `fake confidence'. These are the samples where the samples have high confidence towards the wrong class, this can lead to wrong prediction at the initial layers. This can affect the EEDNN accuracy and needs to be addressed.

\section{Conclusion}
EEDNNs address latency by enabling easier samples to exit at shallower layers, improving both efficiency and accuracy by mitigating overthinking. They also tackle overfitting, vanishing gradients, and distributed inference challenges. While significant progress has been made, ongoing research focuses on optimizing exit criteria, training methods, and addressing issues like overconfidence and prediction errors. This survey highlights key design challenges to inspire further advancements, positioning early-exit techniques as essential tools for future computational systems.
\bibliography{custom}

\end{document}